\documentclass[conference]{IEEEtran}
\IEEEoverridecommandlockouts

\usepackage{cite}
\usepackage[pdftex]{graphicx}
\usepackage{booktabs}
\usepackage{amsmath,amssymb}
\usepackage{tikz}
\usetikzlibrary{calc}
\usepackage{pgffor}
\usepackage[hidelinks]{hyperref}

\def\BibTeX{{\rm B\kern-.05em{\sc i\kern-.025em b}\kern-.08em
    T\kern-.1667em\lower.7ex\hbox{E}\kern-.125emX}}

\begin{document}

\title{A Lightweight Convolutional Neural Network for Real-Time Recognition of Hand-Drawn Geometric Shapes}

\author{
  \IEEEauthorblockN{Shahir Abdullah}
  \IEEEauthorblockA{
    sshahnoor2620025@mscse.uiu.ac.bd
  }
}

\maketitle

\begin{abstract}
Recognizing hand-drawn geometric shapes is a foundational sub-problem of sketch recognition, with applications in education, human-computer interaction, and diagram digitization. This paper presents the design, implementation, and evaluation of a desktop application that recognizes four basic hand-drawn geometric shapes, circle, square, rectangle, and triangle using a compact Convolutional Neural Network (CNN). A dataset of 2{,}000 labeled 28$\times$28-pixel shape images was collected independently and released publicly. The classifier consists of three convolutional blocks (16, 32, and 64 filters) with max-pooling, an in-model data-augmentation stage (random horizontal flip, rotation, and zoom), a dropout-regularized dense layer of 128 units, and a 4-way linear output layer, totaling 97{,}956 trainable parameters. The network is trained with the Adam optimizer on a sparse categorical cross-entropy objective computed directly on logits. On an 80/20 train-validation split, the model achieves 94.80\% training accuracy and 96.01\% validation accuracy with a validation loss of 0.1437. A Tkinter-based graphical interface allows a user to draw a shape with the mouse and receive an immediate class prediction with a confidence score. We situate this system within the broader sketch- and shape-recognition literature, compare its accuracy against related hand-drawn shape classification studies, and discuss the limitations inherent to a small, single-contributor dataset. The complete source code, trained model, and per-class datasets are released publicly to support reproducibility.
\end{abstract}


\section{Introduction}

Sketching is one of the most natural ways for humans to externalize visual and spatial ideas, and geometric primitives. Geometric shapes such as circles, squares, rectangles, and triangles are the building blocks from which more complex hand-drawn diagrams, charts, and figures are composed. Automatically recognizing these primitives from freehand input has practical value for educational software that gives students instant feedback on geometry exercises, for note-digitization pipelines that convert whiteboard sketches into structured diagrams, and for sketch-based user interfaces more broadly.

While humans distinguish a circle from a triangle almost instantly and with near-perfect accuracy, teaching a machine to do the same from raw pixel input is non-trivial: strokes vary in thickness, closure, symmetry, and scale, and classical geometric descriptors are brittle to these variations. Convolutional Neural Networks (CNNs) address this by learning spatial-hierarchical features directly from pixel data, removing the need for hand-engineered shape descriptors, and have become the default approach for small-image classification problems of this kind.

This paper documents a complete, self-contained pipeline for hand-drawn geometric shape recognition: a custom-collected dataset, a compact CNN trained on that dataset, and a working Tkinter desktop application that captures freehand mouse input and returns a real-time shape prediction with a confidence score. The specific contributions of this paper are:

\begin{itemize}
  \item A custom dataset of 2{,}000 labeled, hand-drawn 28$\times$28 images spanning four geometric classes (circle, square, rectangle, triangle), released publicly on Kaggle for reuse by other researchers.
  \item A lightweight CNN architecture, three convolutional blocks with 16, 32, and 64 filters, an in-model data-augmentation stage, and 97{,}956 trainable parameters in total that reaches 96.01\% validation accuracy despite the small dataset size.
  \item A working, open-source graphical desktop application (Tkinter) that demonstrates end-to-end, real-time inference from freehand mouse input, including the exact preprocessing chain (grayscale conversion, resizing, edge enhancement) used at inference time.
\end{itemize}

The remainder of this paper is organized as follows. Section~\ref{sec:related} reviews related work in classical shape recognition, sketch-recognition benchmarks, and CNN-based approaches to hand-drawn shape classification. Section~\ref{sec:dataset} describes the dataset. Section~\ref{sec:system} details the proposed system, including the exact network architecture recovered from the released model checkpoint. Section~\ref{sec:results} reports experimental results. Section~\ref{sec:comparison} compares these results against related studies. Section~\ref{sec:discussion} discusses the findings, Section~\ref{sec:limitations} enumerates the limitations and threats to validity, and Section~\ref{sec:conclusion} concludes with directions for future work.

\section{Related Work}
\label{sec:related}

\subsection{Classical and Geometric Approaches}

Prior to the widespread adoption of learned representations, shape recognition relied on geometric and contour-based descriptors. The Hough transform \cite{duda1972hough} detects parametric primitives such as lines and circles by accumulating votes in a parameter space, and remains a standard tool for detecting well-formed geometric primitives in clean line drawings. Contour-based techniques approximate a shape's outline with polygons or moment invariants and classify based on descriptors such as corner count, aspect ratio, or circularity. These methods can be effective on clean, noise-free input, but are sensitive to stroke gaps, self-intersections, and the irregularities typical of freehand sketching, which motivates learning-based alternatives.

\subsection{Sketch-Recognition Datasets and Benchmarks}

Eitz et al.~\cite{eitz2012humans} conducted the first large-scale study of human sketching behavior, collecting 20{,}000 sketches spanning 250 object categories (the TU-Berlin sketch dataset) and showing that even humans correctly classify unseen sketches only 73\% of the time, while a bag-of-features representation with a multi-class SVM reached 56\% accuracy. This result illustrates that general object-level sketch recognition is intrinsically difficult due to high intra-class variability. At a larger scale, Ha and Eck~\cite{ha2018sketchrnn} introduced the QuickDraw-derived sketch-rnn framework, trained on millions of vector sketches collected through the Quick, Draw! game, and modeled sketches as sequences of pen strokes rather than static raster images, enabling both recognition and generative sketch modeling. These works establish that sketch recognition spans a spectrum from small, well-defined primitive sets (as addressed in this paper) to broad, open-vocabulary object recognition, with corresponding differences in achievable accuracy.

\subsection{CNN-Based Recognition of Hand-Drawn Shapes}

The use of small CNNs on low-resolution, grayscale-like images was established by LeCun et al.~\cite{lecun1998gradient}, whose LeNet architecture on 28$\times$28/32$\times$32 digit images directly motivates the input resolution used in this work and in many subsequent small-image classification systems. Larger-scale CNNs such as AlexNet~\cite{krizhevsky2012imagenet} later demonstrated that depth and parameter count can be scaled substantially for richer natural-image tasks, though such capacity is unnecessary for a four-class geometric primitive problem of the kind addressed here.

Most directly related to the present work, Alam et al.~\cite{alam2025transfer} proposed CnN-RFc, a hybrid CNN and random-forest classifier for hand-drawn mathematical geometric shapes, evaluated on a benchmark of 20{,}000 images across eight classes (circle, kite, parallelogram, square, rectangle, rhombus, trapezoid, triangle), reporting 98\% accuracy with a Light Gradient Boosting Machine variant. Their work confirms that CNN-based feature extraction is well suited to hand-drawn geometric-shape classification and provides a useful accuracy reference point, albeit on a larger and more class-diverse benchmark than the one used here (Section~\ref{sec:comparison}).

Regularization techniques adopted in this work are similarly well established: dropout~\cite{srivastava2014dropout} randomly deactivates units during training to reduce co-adaptation and overfitting, and data augmentation surveyed broadly by Shorten and Khoshgoftaar~\cite{shorten2019survey} synthesizes label-preserving input variations (flips, rotations, zooms) to improve generalization when training data is limited, which is directly relevant given the modest size of the dataset used in this study.

In summary, this work differs from the broader sketch-recognition literature in scope: rather than targeting large, open-vocabulary sketch categories (as in \cite{eitz2012humans,ha2018sketchrnn}), it focuses on a minimal, well-defined set of four geometric primitives, using a deliberately lightweight CNN (under 100{,}000 parameters) suited to real-time CPU inference within a desktop GUI, rather than a server or GPU-class model.

\section{Dataset}
\label{sec:dataset}

\subsection{Data Collection}

No public dataset of hand-drawn geometric shapes suited to this task was readily available at the time of the original project, so a dataset of 2{,}000 images spanning four classes, circle, square, rectangle, and triangle was created from scratch. Shapes were hand-drawn and compressed to a uniform resolution of 28$\times$28 pixels, matching the input resolution conventions established by classic small-image benchmarks such as MNIST~\cite{lecun1998gradient}. The per-class distribution of the 2{,}000 images is not recorded in the original project documentation and is therefore not reported here; we note this as a documentation gap rather than assert a class balance that cannot be verified.

\subsection{Preprocessing and Splitting}

The complete set of 2{,}000 images was split 80/20 into training and validation partitions using Keras' image-preprocessing utilities. Pixel values are rescaled from the $[0, 255]$ range to $[0, 1]$ by a \texttt{Rescaling} layer (scale factor $1/255$) situated at the input of the network itself, and three data-augmentation layers \texttt{RandomFlip} (horizontal), \texttt{RandomRotation} (factor $\pm0.1$), and \texttt{RandomZoom} (height factor $0.1$) are likewise embedded directly in the model graph and are active only during training, so that a single saved checkpoint contains the complete, reproducible preprocessing pipeline.

\subsection{Dataset Availability}

The four class-wise datasets were published on Kaggle~\cite{kaggle_triangles,kaggle_squares,kaggle_rectangles,kaggle_circles} to support reuse by other researchers working on hand-drawn shape or sketch-recognition tasks.

\begin{figure}[t]
  \centering
  \includegraphics[width=\columnwidth]{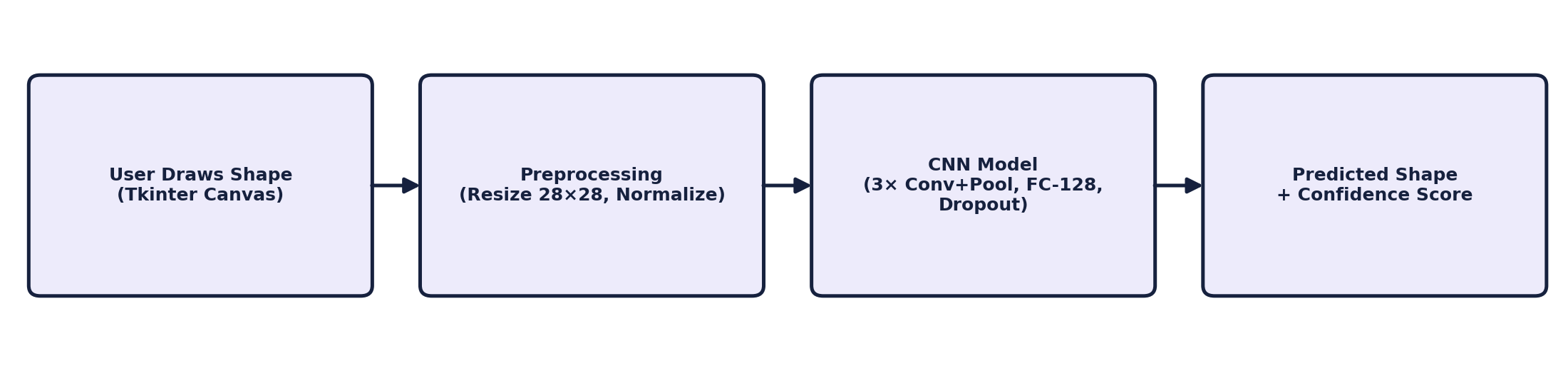}
  \caption{End-to-end system pipeline: canvas input capture, preprocessing, CNN inference, and prediction output.}
  \label{fig:pipeline}
\end{figure}

\section{Proposed System}
\label{sec:system}

\subsection{System Overview}

The system consists of three stages, illustrated in Fig.~\ref{fig:pipeline}: (1) freehand shape capture through a drawing canvas, (2) a fixed preprocessing chain that converts the captured drawing into a normalized 28$\times$28 input tensor, and (3) classification by a trained CNN that outputs one of four class labels together with a softmax confidence score.

\subsection{Drawing Interface and Input Capture}

The front end is a desktop application built with Python's Tkinter toolkit. It provides a 500$\times$500-pixel white canvas on which the user draws freehand with the mouse using a black pen of adjustable width (5~px by default); the in-memory PIL image mirrors every stroke drawn on the visible canvas. When the user requests a prediction, this 500$\times$500 image is converted to a single-channel-plus-alpha (`LA') representation, resized to 28$\times$28 pixels, and sharpened with PIL's \texttt{EDGE\_ENHANCE\_MORE} filter before being written to disk and reloaded through Keras' image-loading utility, which returns it as a 3-channel (RGB) array to match the network's expected $(28, 28, 3)$ input shape.

\subsection{CNN Architecture}

The exact architecture below was recovered directly from the released Keras checkpoint (\texttt{shapedetector\_model\_4b.h5}) rather than reconstructed from memory, and is reported here layer-by-layer for reproducibility. The classifier is a compact CNN consisting of three convolutional blocks (16, 32, and 64 filters respectively, each with 3$\times$3 kernels, same padding, and ReLU activation), each followed by 2$\times$2 max-pooling. The final pooled feature map ($3\times3\times64$) passes through a dropout layer (rate 0.2), is flattened to a 576-dimensional vector, and fed into a fully connected layer of 128 ReLU units before a final 4-unit linear (logit) output layer. Table~\ref{tab:arch} lists every layer, its output shape, and its parameter count.

\begin{table}[t]
  \caption{Network architecture and parameter counts (total: 97{,}956 trainable parameters). *Active only during training.}
  \label{tab:arch}
  \centering
  \begin{tabular}{@{}lcc@{}}
    \toprule
    \textbf{Layer} & \textbf{Output Shape} & \textbf{Params} \\
    \midrule
    Input                              & $28\times28\times3$ & 0 \\
    Data augmentation*                 & $28\times28\times3$ & 0 \\
    Rescaling ($\times 1/255$)         & $28\times28\times3$ & 0 \\
    Conv2D (16, $3\times3$, ReLU)      & $28\times28\times16$ & 448 \\
    MaxPooling2D ($2\times2$)          & $14\times14\times16$ & 0 \\
    Conv2D (32, $3\times3$, ReLU)      & $14\times14\times32$ & 4{,}640 \\
    MaxPooling2D ($2\times2$)          & $7\times7\times32$   & 0 \\
    Conv2D (64, $3\times3$, ReLU)      & $7\times7\times64$   & 18{,}496 \\
    MaxPooling2D ($2\times2$)          & $3\times3\times64$   & 0 \\
    Dropout (rate = 0.2)               & $3\times3\times64$   & 0 \\
    Flatten                            & 576                  & 0 \\
    Dense (128, ReLU)                  & 128                  & 73{,}856 \\
    Dense (4, linear / logits)         & 4                    & 516 \\
    \bottomrule
  \end{tabular}
\end{table}

\subsection{Training Configuration}

The network is trained with the Adam optimizer (learning rate $= 0.001$, $\beta_1 = 0.9$, $\beta_2 = 0.999$, $\varepsilon = 1\times10^{-7}$), minimizing sparse categorical cross-entropy computed directly on the logits (\texttt{from\_logits=True}) for numerical stability, with classification accuracy tracked as the reported metric. These values are taken directly from the training configuration embedded in the released checkpoint. The exact number of training epochs, batch size, and total training time were not recorded in the original project artifacts and are therefore not reported; a fully reproducible re-run would require retraining with the released dataset and code under an explicitly logged configuration.

\section{Experimental Results}
\label{sec:results}

\subsection{Evaluation Protocol}

The model was evaluated on the 20\% validation split held out from the 2{,}000-image dataset (Section~\ref{sec:dataset}-B). No separate, independently held-out test set beyond this validation split is described in the original project documentation; this is noted explicitly as a limitation in Section~\ref{sec:limitations}.

\subsection{Quantitative Results}

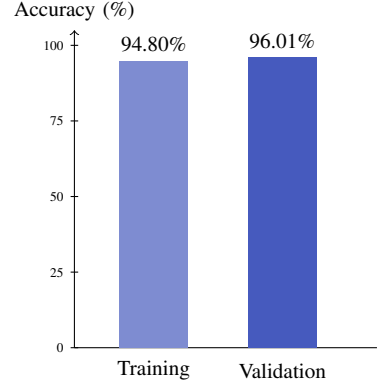
\begin{figure}[t]
  \centering
  \begin{tikzpicture}
    \draw[->] (0,0) -- (0,4.2) node[above] {\footnotesize Accuracy (\%)};
    \draw[-] (0,0) -- (4,0);
    \foreach \y/\lab in {0/0,1/25,2/50,3/75,4/100}{
      \draw (-0.05,\y) -- (0.05,\y) node[left=2pt] {\tiny \lab};
    }
    \definecolor{barblue}{RGB}{70,90,190}
    \fill[barblue!70] (0.6,0) rectangle (1.5,3.792);
    \node at (1.05,3.792) [above] {\footnotesize 94.80\%};
    \node at (1.05,-0.3) {\footnotesize Training};
    \fill[barblue] (2.3,0) rectangle (3.2,3.8404);
    \node at (2.75,3.8404) [above] {\footnotesize 96.01\%};
    \node at (2.75,-0.3) {\footnotesize Validation};
  \end{tikzpicture}
  \caption{Final training vs.\ validation accuracy.}
  \label{fig:accbar}
\end{figure}

Table~\ref{tab:results} summarizes the final training and validation metrics recorded for the model, and Fig.~\ref{fig:accbar} visualizes the training-versus-validation accuracy.

\begin{table}[t]
  \caption{Final training and validation metrics.}
  \label{tab:results}
  \centering
  \begin{tabular}{@{}lc@{}}
    \toprule
    \textbf{Metric} & \textbf{Value} \\
    \midrule
    Training Accuracy      & 94.80\% \\
    Validation Accuracy    & 96.01\% \\
    Validation Loss        & 0.1437 \\
    Trainable Parameters   & 97{,}956 \\
    \bottomrule
  \end{tabular}
\end{table}

\subsection{Model Footprint}

With only 97{,}956 trainable parameters, the network is several orders of magnitude smaller than large-scale image classifiers such as AlexNet~\cite{krizhevsky2012imagenet} (approximately 60 million parameters), making it well suited to real-time CPU-only inference inside a desktop GUI without dedicated graphics hardware consistent with the sub-second prediction latency observed in the accompanying application.

\section{Comparison with Related Work}
\label{sec:comparison}

Table~\ref{tab:comparison} places the reported accuracy of this work alongside two related studies discussed in Section~\ref{sec:related}. The comparison is illustrative rather than a controlled benchmark: the three studies differ substantially in class count, dataset size, and task difficulty, so accuracy figures are not directly comparable and should be read as context rather than a ranked evaluation.

\begin{table}[t]
  \caption{Illustrative comparison with related hand-drawn shape/sketch recognition studies.}
  \label{tab:comparison}
  \centering
  \begin{tabular}{@{}lccc@{}}
    \toprule
    \textbf{Study} & \textbf{Classes} & \textbf{Images} & \textbf{Accuracy} \\
    \midrule
    Eitz et al.~\cite{eitz2012humans}, 2012 & 250 (general) & 20,000 & 56\% / 73\%$^{\dagger}$ \\
    Alam et al.~\cite{alam2025transfer}, 2025 & 8 (math.\ shapes) & 20,000 & 98\% \\
    This work, 2021 & 4 (geometric) & 2,000 & 96.01\% (val.) \\
    \bottomrule
  \end{tabular}
  \\[2pt]
  \raggedright\footnotesize $^{\dagger}$computational / human accuracy, respectively.
\end{table}

Two observations follow. First, the accuracy achieved here is broadly consistent with Alam et al.~\cite{alam2025transfer}, supporting the view that hand-drawn geometric-primitive classification is a comparatively tractable sub-problem of the broader sketch-recognition space addressed in~\cite{eitz2012humans}, where general object-category sketches are recognized correctly only 56-73\% of the time even by humans and dedicated computer-vision pipelines. Second, this work reaches its accuracy with roughly an order of magnitude less training data and a far smaller network than~\cite{alam2025transfer}, suggesting that for a minimal four-class geometric vocabulary, a lightweight CNN is sufficient without requiring a hybrid ensemble classifier.

\section{Discussion}
\label{sec:discussion}

The validation accuracy (96.01\%) exceeding the training accuracy (94.80\%) is consistent with the regularization effect of the embedded dropout and data-augmentation layers (Section~\ref{sec:system}-C, D), which make the training-time task deliberately harder than plain inference on the (non-augmented) validation set a common and expected pattern when augmentation is applied only during training.

More broadly, the results show that a compact, sub-100{,}000-parameter CNN can achieve strong classification performance on a constrained, four-class shape-recognition problem even when trained on a modest, self-collected dataset of 2{,}000 images, without requiring transfer learning or a large pretrained backbone. This supports the use of small, task-specific CNNs for narrowly scoped sketch-recognition problems, in contrast to the larger architectures and datasets typically used for open-vocabulary sketch recognition (Section~\ref{sec:related}-B).

\section{Limitations and Threats to Validity}
\label{sec:limitations}

\begin{itemize}
  \item \textbf{Single-contributor handwriting:} all 2{,}000 images were hand-drawn by one person, which may limit generalization to users with substantially different drawing styles, stroke widths, or input devices (e.g., touchscreen or stylus input, which was not evaluated).
  \item \textbf{No independent test set:} the original project documentation describes only an 80/20 train/validation split, with no separate held-out test set used solely for final reporting; the validation accuracy reported here may therefore be optimistic relative to unseen, out-of-distribution input.
  \item \textbf{Undocumented training configuration:} epoch count, batch size, and total training time are not recorded in the released artifacts, limiting exact reproducibility of the training run (though the architecture and optimizer settings are fully recovered and reported in Section~\ref{sec:system}).
  \item \textbf{Narrow class scope:} the classifier distinguishes only four convex/simple primitives and has not been evaluated on additional polygons, overlapping shapes, incomplete strokes, or a rejection (`not a shape') class.
  \item \textbf{Unreported per-class metrics:} only aggregate accuracy and loss are available from the original project artifacts; a per-class confusion matrix was not recorded and is not fabricated here.
\end{itemize}

\section{Conclusion and Future Work}
\label{sec:conclusion}

This paper presented the design, exact architecture, and evaluation of a lightweight CNN-based system for recognizing hand-drawn geometric shapes, together with a working Tkinter desktop interface for real-time use. Recovered directly from the released model checkpoint, the network uses 97{,}956 trainable parameters across three convolutional blocks and a dense classification head, trained with Adam on sparse categorical cross-entropy, and reaches 96.01\% validation accuracy on a self-collected, four-class, 2{,}000-image dataset.

Future work includes: expanding the label set to additional polygons (e.g., pentagons, hexagons, stars) and a negative/rejection class for non-shape scribbles; collecting drawings from multiple contributors, including touchscreen and stylus input, to improve generalization; introducing an explicit, held-out test set distinct from the validation split used for model selection; logging and publishing the full training configuration (epochs, batch size, hardware, wall-clock time) for exact reproducibility; and benchmarking lightweight alternative architectures (e.g., depthwise-separable convolutions) for further reductions in parameter count and inference latency on constrained or embedded devices.

\end{document}